# Two-Stream Multi-Channel Convolutional Neural Network (TM-CNN) for Multi-Lane Traffic Speed Prediction Considering Traffic Volume Impact

Ruimin Ke[1], Wan Li[2], Zhiyong Cui[1], Yinhai Wang[1,*]

[1] Smart Transportation Applications and Research (STAR) Lab, Department of Civil and Environmental Engineering, University of Washington
[2] The intelligent Urban Transportation Systems (iUTS) Lab, Department of Civil and Environmental Engineering, University of Washington

[*] yinhai@uw.edu

**Abstract:** Traffic speed prediction is a critically important component of intelligent transportation systems (ITS). Recently, with the rapid development of deep learning and transportation data science, a growing body of new traffic speed prediction models have been designed, which achieved high accuracy and large-scale prediction. However, existing studies have two major limitations. First, they predict aggregated traffic speed rather than lane-level traffic speed; second, most studies ignore the impact of other traffic flow parameters in speed prediction. To address these issues, we propose a two-stream multi-channel convolutional neural network (TM-CNN) model for multi-lane traffic speed prediction considering traffic volume impact. In this model, we first introduce a new data conversion method that converts raw traffic speed data and volume data into spatial-temporal multi-channel matrices. Then we carefully design a two-stream deep neural network to effectively learn the features and correlations between individual lanes, in the spatial-temporal dimensions, and between speed and volume. Accordingly, a new loss function that considers the volume impact in speed prediction is developed. A case study using one-year data validates the TM-CNN model and demonstrates its superiority. This paper contributes to two research areas: (1) traffic speed prediction, and (2) multi-lane traffic flow study.

## 1. Introduction

Traffic speed prediction is one of the most crucial components of intelligent transportation systems (ITS). It can benefit both traffic agencies and travelers by contributing to key applications such as variable speed limit control and route guidance. Although traffic speed prediction has a long history that can be dated back to several decades ago, traditional traffic speed prediction methods are unable to precisely capture the high dimensional and nonlinear characteristics of traffic flow due to the lack of either the computational ability or amount of data [1, 2]. In recent years, with the emerging trends in artificial intelligence and transportation data science, a growing body of research has been conducted in this field.

The typical traffic speed prediction problem is to predict traffic speed at a future time using given historical traffic data. Traditionally, time series methods such as autoregressive integrated moving average (ARIMA) and conventional machine learning models such as support vector regression (SVR) are widely applied to traffic prediction [3–12]. Later on, with the tremendous success of deep learning in many fields [13–15], researchers started to explore the possibility of deep learning for traffic speed prediction and then developed a couple of deep learning models that significantly outperformed the conventional models [16–21]. For example, Ma et al. implemented long short-term memory neural network (LSTM NN) for the first time in traffic speed prediction. Their work suggested that LSTM NN received the best performance over previous methods [19]. Tang et al. designed an improved fuzzy neural network (FNN) for traffic speed prediction. This model considered the periodic characteristics of traffic flow and achieved state-of-the-art performance [20]. Although these pioneering studies still focus on relatively small-scale prediction at individual locations, they have greatly inspired the explorations of more advanced deep-learning-based traffic speed prediction methods.

Recently, substantial research has focused on extending the traffic speed prediction problem from individual roadway locations to traffic networks by designing new deep neural networks that integrate physical roadway structures [22–30]. Ma et al. developed a convolutional neural network (CNN) model that can capture spatial correlations between adjacent roadway segments and temporal correlations between adjacent times in a 2D spatial-temporal matrix [22]. Yao et al. proposed a deep learning architecture named spatial-temporal dynamic network (STDN) that incorporated CNN, LSTM, and a periodically shifted attention mechanism to address the issues on dynamic dependency and shifting of long-term periodic dependency [29]. Cui et al. devised a high-order graph convolutional LSTM (HGC-LSTM) to model the dynamics of the traffic speed and acquire the spatial dependencies within the traffic network. This group of studies considers both spatial dependencies and temporal dynamics of traffic flow in deep learning models, thereby enables effective learning and accurate speed prediction for network-scale traffic.

Despite the achievements mentioned above in traffic speed prediction, the existing studies have two major limitations. First, they predict aggregated traffic speed rather than lane-level traffic speed. At every data collection unit of roadways, they implicitly assume no traffic pattern difference between different lanes. In some studies, this is due to the unavailability of lane-level traffic data; in others where the lane-level data are available, the speeds are still often aggregated for simplifying the model complexity. However, since a long time ago, research has revealed that traffic flows on different lanes show different yet correlated patterns [31–39]. For instance, Daganzo et al. studied a "reverse lambda" pattern in their work [37]. This pattern shows as consistently high flows on freeway median lanes, but it has not been reported for the shoulder lanes. It is also observed that for either two-lane or three-lane freeway segments, there are certain volume-density distributions for individual lanes [34]. As the increasing need for lane-based traffic operations such as carpool lane tolling and reversable lane control in modern transportation systems, this issue can no longer be ignored.

The second limitation is that most existing studies ignore other traffic flow parameters in speed prediction tasks. In traffic flow theory, there are correlations among traffic flow speed, volume, and occupancy [40]. Without the integration of volume or occupancy into speed prediction, the hidden traffic flow patterns may not be fully captured and learned, which can lead to reduced prediction accuracy [41]. An intuitive example is that: In free-flow conditions, a larger-volume traffic stream tends to be more sensitive to perturbances than smaller-volume traffic stream. Therefore, the speed of the larger-volume traffic stream is more likely to decrease in a future time. However, without the volume or occupancy data, it is hard to model the hidden traffic flow patterns.

To address these challenges, we propose a two-stream multi-channel convolutional neural network (TM-CNN) for multi-lane traffic speed prediction with the consideration of traffic volume impact. In the proposed model, we develop a data conversion method to convert both the multi-lane speed data and multi-lane volume data into multi-channel spatial-temporal matrices. We design a CNN architecture with two streams, where one takes the multi-channel speed matrix as input and another takes the multi-channel volume matrix as input. A fusion method is further implemented for the two streams. Specifically, convolutional layers learn the two matrices to capture traffic flow features in three dimensions: the spatial dimension, the temporal dimension, and the lane dimension. Then, the output tensors of the two streams will be flattened and concatenated into one speed-volume vector, and this vector will be learned by the fully connected (FC) layers. Accordingly, a new loss function is devised considering the volume impact in the speed prediction task.

The proposed TM-CNN model is validated using one-year loop detector data on a major freeway in the Seattle area. The comprehensive comparisons and analyses demonstrate the strength and effectiveness of our model. This paper contributes to two transportation research areas. First, it contributes to the traffic speed prediction area by adding a new deep neural network model to the existing literature. Second, it pushes off the boundary of knowledge in the multi-lane traffic flow study area by developing a method for the learning and speed prediction of multi-lane traffic. In summary, the contribution of this paper is fourfold:

(1) We introduce a new data conversion method to convert the multi-lane traffic speed data and volume data into spatial-temporal multi-channel matrices. The converted data matrices are organized as the inputs to the deep neural network.
(2) We design a two-stream CNN architecture for multi-lane traffic speed prediction. The convolutional layers extract the correlations between lanes and spatial-temporal features in the multi-channel data matrices. It also concatenates the outputs of the two convolutional-layer streams and learns a speed-volume feature vector.
(3) We propose a new loss function for the deep learning model. It is the sum of a speed term and a weighted volume term. By appropriately setting the weight, the volume term improves the learning ability of the model and helps prevent overfitting.
(4) Traditional studies on multi-lane traffic flow mostly focus on the mathematically modeling and behavior description of multi-lane traffic. This study is among the first efforts to apply deep learning methods for multi-lane traffic pattern mining and prediction.

## 2. Methodology

### 2.1 *Modeling multi-lane traffic as multi-channel matrices*

The first step of our methodology is modeling the multi-lane traffic flow as multi-channel matrices. We propose a data conversion method to convert the raw data into spatial-temporal multi-channel matrices, in which traffic on every individual lane is added to the matrices as a separate channel. This modeling idea comes from CNN's superiority to capture features in multi-channel RGB images. In RGB images, each color channel has correlations yet differences with the other two. This is similar to traffic flows on different lanes where correlations and differences both exist [32, 37]. Thus, averaging traffic flow parameters at a certain milepost and timestamp is like doing a weighted average of the RGB values to get the grayscale value. In this sense, previous methods for traffic speed prediction are designed for "grayscale images" (spatial-temporal prediction for averaged speed) or even just a single image column (speed prediction for an individual location). In this study, the proposed model manages to handle lane-level traffic information by formulating the data inputs as "RGB images."

In this paper, loop detector data is used due to the fact it collects different types of traffic flow data on individual lanes. That is being said, though loop detector is a relatively traditional traffic detector, it provides lane-level traffic speed, volume, and occupancy data which many other detectors do not [42–44]. For example, probe vehicle data are widely used nowadays, but besides a small sample of traffic speeds and trajectories, most of them are unable to collect lane-level data or volume data.

This data conversion method diagram is shown in Figure 1. There are loop detectors installed at $k$ different mileposts along this segment, and the past $n$ time steps are considered in the prediction task. We denote the number of lanes as $c$. Without loss of generality, it is assumed that the number of lanes is three in Figure 1 for the sake of illustration. Single-lane traffic would be represented by two $k \times n$ spatial-temporal 2D matrices, where one is for speed and another for volume. We denote them as $I_u$ for speed and $I_q$ for volume. We define the speed value and volume value to be $u_{ilt}$ and $q_{ilt}$ respectively for a detector at milepost $i$ ($i$

= 1,2,...,k) and lane l (l = 1,2,...,c) at time t (t = 1,2,...,n). Note that each $u_{ilt}$ or $q_{ilt}$ is normalized to between 0 and 1 using min-max normalization since speed and volume have different value ranges. Hence, in the speed and volume matrices with the size $k \times n \times c$, we construct the matrices using Eq. (1) and Eq. (2),

$$I_u(i,t) = (u_{i1t}, u_{i2t}, ..., u_{ict}) \quad (1)$$
$$I_q(i,t) = (q_{i1t}, q_{i2t}, ..., q_{ict}) \quad (2)$$

where $i$ and $t$ are the row index and column index of a spatial-temporal matrix, representing the milepost and the timestamp, respectively. $I_u(i,t)$ and $I_q(i,t)$ denote the multi-channel pixel values of the speed and the volume. The number of channels correspond to the number of lanes $c$. Each element in the 2D multi-channel matrices is a $c$-unit vector representing $c$ lanes' traffic speeds or volumes at a given milepost $i$ and time $t$. In the three-lane example in Figure 1, the spatial-temporal matrices have three channels. Mathematically, the spatial-temporal multi-channel matrices for traffic speed ($X_u$) and volume ($X_q$) can be denoted as

$$X_u = \begin{bmatrix} I_u(1,1) & I_u(1,2) & ... & I_u(1,n) \\ I_u(2,1) & I_u(2,2) & ... & I_u(2,n) \\ \vdots & \vdots & & \vdots \\ I_u(k,1) & I_u(k,2) & ... & I_u(k,n) \end{bmatrix} \quad (3)$$

$$X_q = \begin{bmatrix} I_q(1,1) & I_q(1,2) & ... & I_q(1,n) \\ I_q(2,1) & I_q(2,2) & ... & I_q(2,n) \\ \vdots & \vdots & & \vdots \\ I_q(k,1) & I_q(k,2) & ... & I_q(k,n) \end{bmatrix} \quad (4)$$

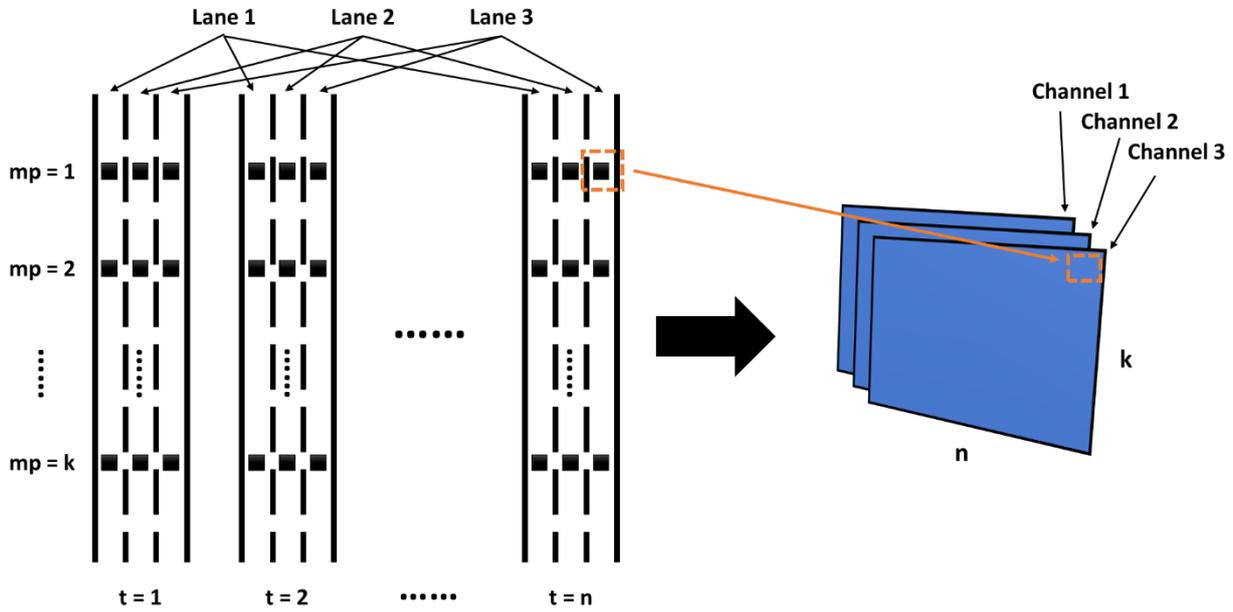

*Fig. 1* The data input modeling process of converting the multi-lane traffic flow raw data to the multi-channel spatial-temporal matrix

## 2.2 Convolution for feature extraction

The CNN has demonstrated a promising performance in image classification and many other applications due to its locally-connected layers and the better ability than other neural networks to capture local features. In transportation, traffic stream, as well as disturbance to traffic stream, moves along the spatial axis and the temporal axis. Thus, applying CNN to the spatial-temporal traffic image manages to capture local features in both spatial and temporal dimensions. The fundamental operation in the feature extraction process of CNN is convolution. With the re-organized input as a multi-channel matrix $X$ ($X$ could be $X_u$ or $X_q$), the basic unit of a convolution operation is shown in Figure 2. On the left most of the figure, it is the input spatial-temporal matrix or image $X$. Every channel of the input matrix is a 2D spatial-temporal matrix representing the traffic flow pattern on the corresponding lane. On the top of the left-most column, channel #1 displays the traffic pattern of lane #1; and on the bottom, the pattern of lane #c is presented. The symbol "*" denotes the convolution operation in Figure 2. Since our input is a multi-channel image, the convolution filters are also multi-channel. In the figure, a 3 × 3 × c filter is drawn, while the size of the filter can be changed in practice. The values inside the cells of a filter are weights of the CNN, which are automatically modified during the training process. The final weights are able to extract the most salient features in the multi-channel image. The convolution operation outputs a feature map for each channel, and they are summed up to be the extracted feature map of this convolution filter in the current convolutional layer. With multiple filters operated on the same input image, a multi-channel feature map will be constructed, and serves as the input to the next layer.

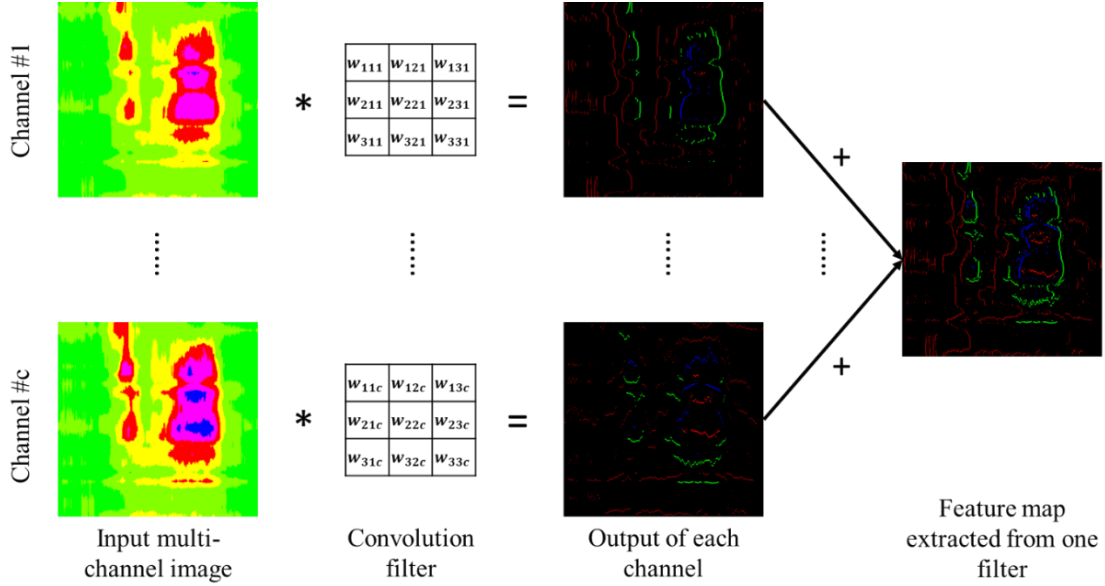

*Fig. 2* The convolution operation to extract features from the multi-channel spatial-temporal traffic flow matrices

### 2.3 The TM-CNN for speed prediction

In order to learn the multi-lane traffic flow patterns and predict traffic speeds, a CNN structure is designed (see Figure 3). Compared to a standard CNN, the proposed CNN architecture is modified in the following aspects: (1) The network inputs are different, that is, the input image is a spatial-temporal image built by traffic sensor data, and it has multiple channels which represent the lanes of a corridor. Moreover, the pixels values' range is different from a normal image. For a normal image, it is 0 to 255; however, here it ranges from 0 to either the highest speed (often the speed limit) or the highest volume (often the capacity). (2) The neural network has two streams of convolutional layers, which are for processing the speeds and volumes. But most CNN has only one stream of convolutional layers. The purpose of having two streams of convolutional layers is to integrate both speed information and volume information into the model so that the network can learn the traffic patterns better than only learning speed. To combine the two streams, a fusion operation that flattens and concatenates the outputs of the two streams are implemented between the convolutional layers and the FC layers. The fusion operation is chosen to be concatenation instead of addition or multiplying because the concatenation operation is more flexible for us to modify each stream's structures. In other words, the concatenation fusion method allows the two streams of convolutional layers to have different structures. (3) The extracted features have unique meanings and are different from image classifications or most other tasks. The extracted features here are relations among road segments, time series, adjacent lanes, and between traffic flow speeds and volumes. (4) The output is different, i.e., our output is a vector of traffic speeds of multiple locations at a future time rather than a single category label or some bounding boxes' coordinates. The output itself is part of the input for another prediction, while this is not the case for most other CNN's.

(5) Different from most CNN's, our CNN does not have a pooling layer. The main reason for not inserting pooling layers in between convolutional layers is that our input images are much smaller than regular images for image classification or object detection [45, 46]. Regular input images to a CNN usually have hundreds of columns and rows while the spatial-temporal images for roadway traffic are not that large. In this research and many existing traffic prediction studies, the time resolution of the data is five minutes, which means even using two-hour data for prediction there are only 24 time steps. Thus, we do not risk losing information by pooling. (6) The loss function is devised to contain both speed and volume information. For traditional image classification CNN's, the loss function is the cross-entropy loss. And for traffic speed prediction tasks, the loss function is commonly the Mean Squared Error (MSE) function with only speed values. However, in this research, we add a new term in the loss function to incorporate the volume information. We denote the ground truth speed vector and volume vector as $Y_u$ and $Y_q$, and the predicted speed vector and volume vector as $\hat{Y}_u$ and $\hat{Y}_q$. Note that $Y_u$, $Y_q$, $\hat{Y}_u$, and $\hat{Y}_q$ are all normalized between 0 and 1. The loss function $L$ is defined in Eq. (5) by summing up the MSEs of speed and volume. The volume term $\lambda||\hat{Y}_q - Y_q||_2^2$ is added to the loss function for reducing the probability of overfitting by helping the model better understand the essential traffic patterns. This design improves the speed prediction accuracy on test dataset with proper settings of λ. Our suggested value of λ is between 0 and 1 considering that the volume term that deals with overfitting should still have a lower impact than the speed term on speed prediction problems.

$$L = ||\hat{Y}_u - Y_u||_2^2 + \lambda||\hat{Y}_q - Y_q||_2^2 \qquad (5)$$

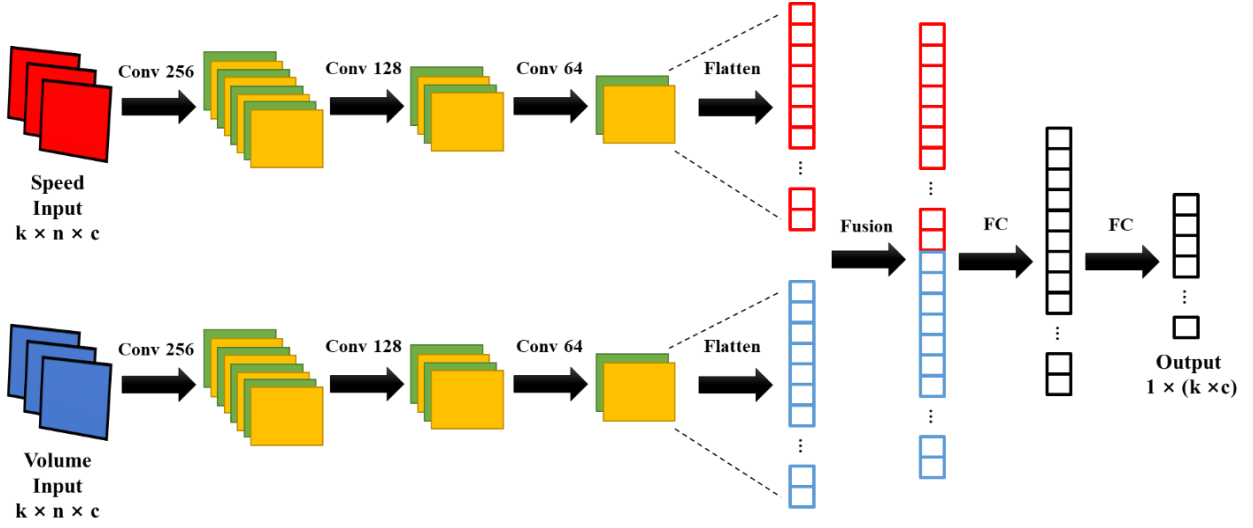

*Fig. 3 The proposed two-stream multi-channel convolutional neural network (TM-CNN) architecture*

In the proposed TM-CNN, the inputs are our multi-channel matrices $X_u$ and $X_q$ with the dimension of $k \times n \times c$. The filter size is all $2 \times 2 \times c$ in order to better capture the correlations between each pair of adjacent loops as well as adjacent times. The number of filters for each convolutional layer is chosen based on experience and the consideration to balance efficiency and accuracy. The last convolutional layer in each of the two streams is flattened and connected to a fully-connected (FC) layer. This FC layer is fully connected with the output layer as well. The length of the output vector $\hat{Y}_u$ is $1 \times (k \times c)$, since the prediction is for one future step. All activations except the output layer use Relu function. The output layer has a linear activation function, which is adopted for regression tasks. Eq. (6) and Eq. (7) describe the derivations mathematically from inputs to the outputs of the last convolutional layers,

$$\hat{Y}_u^{conv} = \varphi\{W_{u3} * \varphi[W_{u2} * \varphi(W_{u1} * X_u + b_{u1}) + b_{u2}] + b_{u3}\} \quad (6)$$

$$\hat{Y}_q^{conv} = \varphi\{W_{q3} * \varphi[W_{q2} * \varphi(W_{q1} * X_q + b_{q1}) + b_{q2}] + b_{q3}\} \quad (7)$$

where $\hat{Y}_u^{conv}$ and $\hat{Y}_q^{conv}$ are the intermediate speed and volume outputs of the CNN in between the last convolutional layers and the flatten layers, $W_{ui}$ and $W_{qi}$ ($i = 1,2,3$) are the weights for the convolutions, $b_{ui}$ and $b_{qi}$ ($i = 1,2,3$) are the biases, and $\varphi(\cdot)$ is the Relu activation function. After getting these two intermediate outputs, we flatten them and fuse them into one vector, and then further learn the relations between the volume feature map and the speed feature map using FC layers. As aforementioned, we choose concatenation as the fusion function for the two flattened intermediate outputs to allows the customization of different neural network designs of the two streams. Customized streams could result in two intermediate outputs of different dimensions. While concatenation would still successfully fuse the two outputs together and support the learning of speed-volume relationships by the FC layers, most other fusion operations require the two vectors to have the same length. This fusion process is mathematically represented in Eq. (8) as follows,

$$\hat{Y}_u, \hat{Y}_q = W_5 \times \varphi\left[W_4 \times Conc\left(F(\hat{Y}_u^{conv}), F(\hat{Y}_q^{conv})\right) + b_4\right] + b_5 \quad (8)$$

where $W_4$ and $W_5$ are the weights for the two FC layers, $b_4$ and $b_5$ are the biases, $F(\cdot)$ is the flatten function, and $Conc(\cdot)$ is the concatenation function.

## 3. Case Study
### 3.1 *Data description*

In this paper, one-year loop data from January 1st, 2016 to December 31st, 2016 for a four-lane freeway corridor in Seattle is used for validation. Seattle is currently a city with top ten busiest traffic in the United States. And this study freeway segment is one of the busiest corridors in Seattle. It starts from milepost-170 to milepost-165 of Interstate-5 (I5) freeway southbound, connecting the University of Washington to Downtown Seattle. There are 40 loop detectors on this corridor. They collect speed, volume, and occupancy traffic data. In our study, we use speed and volume for speed prediction. The reason we do not include occupancy in our model is that adding occupancy increases the training time and complexity yet does not improve the prediction accuracy. This can be explained by traffic flow theory: In most cases, if two of the three traffic parameters are known, the third one can be estimated. Hence, essentially, using three of them does not add more information to the prediction model. The data is downloaded from a traffic big data platform named Digital Roadway Interactive Visualization and Evaluation Network (DRIVE Net) [47], where the data is aggregated to every 5 minutes. Based on our data conversion method, speed and volume are each converted to a four-channel matrix. The data conversion of the one-year data generates about 105,000 data samples for model validation.

### 3.2 *Model implementation*

The proposed model is implemented in Keras deep learning library using TensorFlow backend on an Nvidia GTX 1080 GPU. The implemented model architecture in the case study is shown in Figure 4. This architecture figure is automatically generated by Keras after the model design. It displays the overall model structure as well as the dimensions of the inputs and outputs of all layer. Each input data sample is organized into the dimension of 10×8×4, where 10 is the number of detectors along the freeway segment on each lane, 8 is the number of time steps used for learning and prediction, and 4 is the number of channels (lanes). Here we choose 8 as the number of time steps because we observe that 8 is large enough (40 minutes) to ensure the model adequately captures the past traffic patterns while we target using short past time steps for efficient training. In this model, three convolutional layers are added sequentially to each stream. We select three as the number of hidden convolutional layers based on a trial-and-error process, during which we observe that three convolutional layers constantly outperform just having one of two convolutional layers, yet little improvement is observed with more than three of them. Two dropout layers are added to the model to reduce the probability of overfitting in the training process. One dropout layer is inserted between the last convolutional layer and the concatenation layer, with a dropout ratio 0.5; another is inserted between the FC layer (which is shown as a dense layer in Keras) and the output layer with a dropout ratio 0.25.

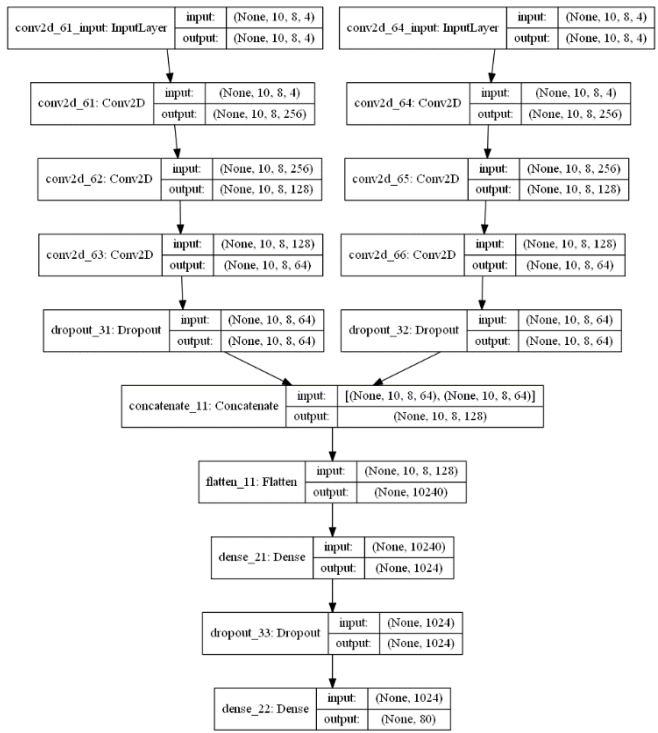

*Fig. 4 The model architecture diagram of the TM-CNN implemented in the case study*

In the model validation process, we split the dataset to 80,000 samples for training and 25,000 samples for testing. In addition to the model architecture parameters, several other hyper-parameters need to be tuned in the training process. For example, the optimizer in our model training is RMSprop given its faster convergence rate than other optimizers. One key hyper-parameter that often influences deep learning models' performances is the learning rate, which determines how much the weights are adjusted with respect to the loss function gradient. It impacts both model accuracy and training speed. In our case, we examined different learning rates ranging from 0.01 to 0.00001, in which we found that the learning rate around 0.0001 generated the best model accuracy. As shown in the first plot in Figure 5, training loss curves are plotted for different learning rates in the first 50 epochs. It can be observed that when the learning rate is smaller than 0.0001, smaller learning rate generates a smaller loss. However, when the learning rate is larger than 0.0001, the model loss starts becoming larger again, and at the same time, the model training takes longer time. Thus, we picked up 0.0001 as the learning rate for our model.

Another critical parameter of the proposed model is the $\lambda$ in the loss function. It determines the impact of traffic volume on the speed prediction. As aforementioned, our suggested value of $\lambda$ is between 0 and 1 with the consideration that the volume term should have a lower impact than the speed term. Therefore, we tested ten values of $\lambda$ from 0 to 0.9 with an interval of 0.1. The curve of speed prediction accuracy on the test dataset with respect to $\lambda$ is shown as the second plot in Figure 5. Compared to no volume term in the loss function, the speed prediction accuracy improves when $\lambda = 0.1$, which implies that the loss function design is effective. The model accuracy starts to decrease as $\lambda$ getting larger from 0.1. This interesting finding indicates that: On the one hand, the volume term does have impact on the speed prediction accuracy; on the other hand, the impact of the volume term should exist but not too large. This observation is reasonable: Firstly, according to traffic flow theory, two traffic flow parameters can better determine the actual traffic flow status than just one parameter; secondly, since volume has more randomness and variation than speed in short term, large impact of volume could increase the uncertainty in speed prediction.

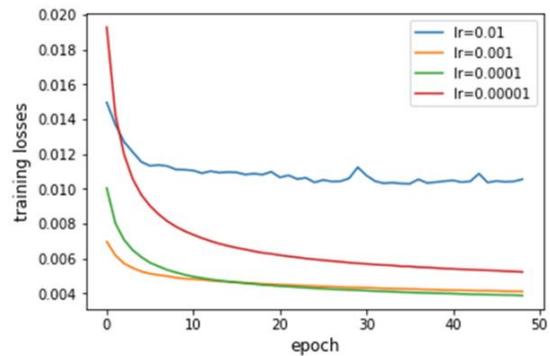

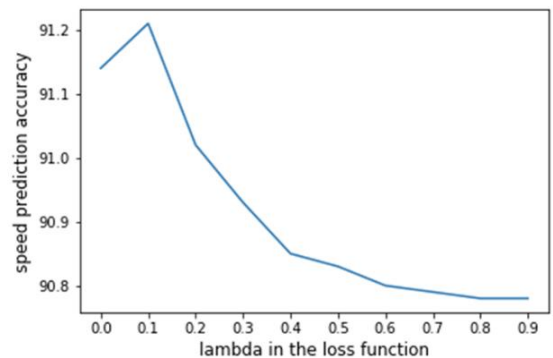

*Fig. 5 Model parameter tuning for learning rate and the $\lambda$ in the loss function*

### 3.3 Results and comparison

In order to demonstrate the superiority of the proposed model, we conducted two evaluations. On the one hand, we compared it with five baseline models, on the other hand, we visualized both the ground truth speeds and the predicted speeds in the formats of spatial-temporal heat map and single-detector speed plot. ARIMA is one of the pioneering methods for traffic prediction; SVR is a popular model in the field before the large-scale applications of deep learning in traffic prediction; ANN is the traditional fully-connected artificial neural network, which often serves as one baseline; LSTM, a specific type of recurrent neural network, is the most widely-used model in recent years for traffic prediction. We also compared the proposed two-stream CNN with a single-stream CNN, which merely contains a speed stream in the network structure and no volume term in the loss function. The models were all finetuned to have their best performances. There are many other speed prediction models, some of which could be more advanced than the baseline models for specific tasks. However, considering most of the existing models are not designed for multi-lane traffic pattern prediction, modifying them to predict multi-lane traffic just for comparison purpose could downgrade their capability and is also not meaningful at this point.

Table 1 shows the accuracies and comparison results. For each model, three different prediction time steps, i.e., 5 mins, 10 mins, and 15 mins are examined. In general, the shorter the prediction time step is, the higher the accuracy. This is consistent with most previous studies. It can be seen that the proposed two-stream multi-channel CNN has the best prediction accuracy over the baseline models in all three cases. The single-stream CNN in general beats the other four baseline models, while has a lower accuracy than the two-stream CNN. Also, it can be observed that with the increase of time step, the prediction accuracy differences between TM-CNN and other models generally become larger. These comparisons show three strengths of the proposed model: First, the conversion of raw traffic data to the multi-channel matrix indeed improves the learning and prediction ability by better capturing spatial-temporal correlations between adjacent lanes, mileposts, and times. Second, the fusion of volume and speed further enhances the learning ability and model accuracy. Third, compared to the baseline methods, the TM-CNN demonstrates a better performance overall and its superiority in relatively longer-term speed prediction.

Figure 6 displays the heat maps of the ground truth speeds (upper) and the predicted speeds (lower) for every lane from 6 am to 8 pm on a day. The prediction time step is 5 minutes in this visualization. The horizontal axis is the index of time. The vertical axis is the index for loop detectors, where loops 0-9 are on lane #1 (the shoulder lane), and loops 30-39 is on lane #4 (the median lane). Figure 7 shows the speed prediction curves (the orange curves) and the ground truth curves (the blue curves) for single loops on every lane at milepost 166.4 for 24 hours. Three observations can be summarized based on the visualizations in Figure 6 and Figure 7: First, the proposed model achieves excellent learning and prediction performances in different traffic conditions (free flow and congestion). Second, the proposed model can learn and capture similar trends yet unique patterns of the traffic flow speed on all individual lanes. Third, the predicted speed values are smoother than the ground truth values. This is due to the variation and noise in real-world traffic flow and traffic data collection. The smoothness of the prediction actually demonstrates the ability of the proposed model to capture the general trends of traffic flow and its robustness to noises.

**Table 1** Accuracy comparison with baseline methods

| | Prediction time steps | | |
|---|---|---|---|
| | 1 (5 mins) | 2 (10 mins) | 3 (15 mins) |
| ARIMA | 83.13% | 81.06% | 78.35% |
| SVR | 82.66% | 80.90% | 78.47% |
| ANN | 87.74% | 85.79% | 83.90% |
| LSTM | 88.46% | 86.78% | 84.65% |
| Single-stream CNN | 90.83% | 89.25% | 86.94% |
| **TM-CNN** | **91.21%** | **90.06%** | **88.15%** |

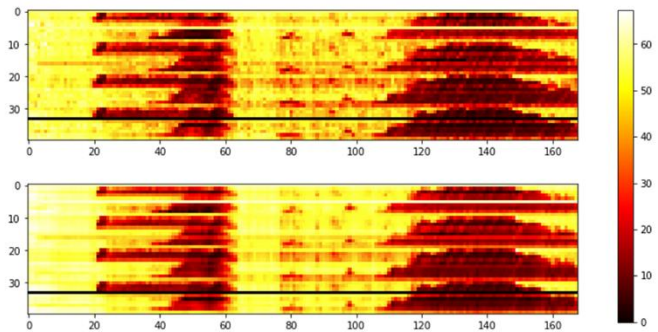

*Fig. 6 Heat maps showing the ground truth speeds (upper) and our predicted speeds (lower) for all four lanes from 6 am to 8 pm on a day*

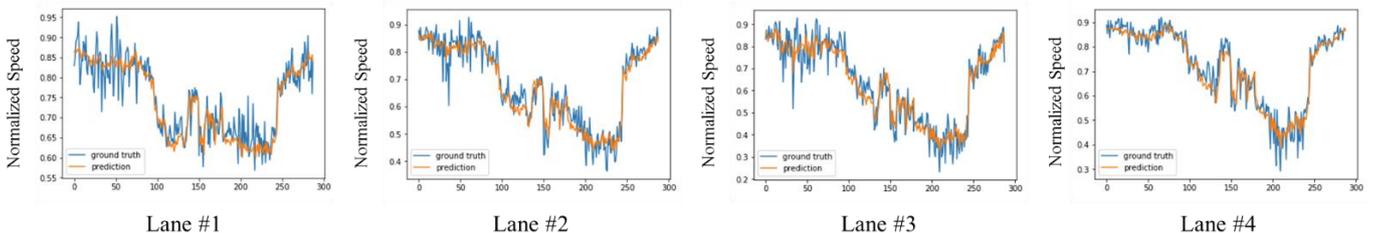

*Fig. 7 The predicted speeds and ground truths at milepost 166.4 for all lanes in 24 hours*

## 4. Conclusion and Future Work

In this paper, we proposed a novel deep learning model called TM-CNN for multi-lane traffic speed prediction. Several new components were carefully designed and incorporated in the model to enable the effective learning of multi-lane traffic flow characteristics and the accurate prediction of multi-lane speeds. The new components included a raw data conversion method, a two-stream multi-channel convolutional neural network architecture, and a new loss function.

Some interesting findings and recommendations can be concluded: (1) Experimental results demonstrate that the TM-CNN can learn and capture the traffic patterns in different traffic conditions and individual lanes. (2) Comparisons with the baseline models show that the TM-CNN achieves superior prediction accuracy and robustness over ARIMA, SVR, ANN, LSTM, and single-stream CNN. (3) The learning rate in the training process and the weight of the volume term in the loss function are critical hyper-parameters in this model. (4) For multi-lane traffic learning and prediction, we suggest converting traffic flow data into multi-channel matrices using the proposed data conversion method. (5) We suggest incorporating traffic volume data into both the neural network architecture and the loss function for speed prediction tasks.

Future work will be carried out in two directions. First, this study conducted an initial experiment on a relatively small-scale dataset for the purpose of validating the model performance on predicting multi-lane traffic. In future studies, we will finetune and test the model for network-scale multi-lane traffic speed prediction. The second future direction is to modify the model structure to integrate ramp detectors data into speed prediction. By doing this, we aim to further improve the learning ability and prediction accuracy of the model.